# LOCALIZATION AND CLASSIFICATION OF PARASITIC EGGS IN MICROSCPIC IMAGES USING AN EFFICIENTDET DETECTOR

*Nouar AlDahoul [1*], Hezerul Abdul Karim [1], Shaira Limson Kee [2], Myles Joshua Toledo Tan [2,3]*

[1]Faculty of Engineering, Multimedia University, Cyberjaya, Malaysia
[2]Department of Natural Sciences, University of St. La Salle, Bacolod City, 6100, Philippines
[3]Department of Chemical Engineering, University of St. La Salle, Bacolod City, 6100, Philippines
[*]nouar.aldahoul@live.iium.edu.my

**Abstract**

IPIs caused by protozoan and helminth parasites are among the most common infections in humans in LMICs. They are regarded as a severe public health concern, as they cause a wide array of potentially detrimental health conditions. Researchers have been developing pattern recognition techniques for the automatic identification of parasite eggs in microscopic images. Existing solutions still need improvements to reduce diagnostic errors and generate fast, efficient, and accurate results. Our paper addresses this and proposes a multi-modal learning detector to localize parasitic eggs and categorize them into 11 categories. The experiments were conducted on the novel Chula-ParasiteEgg-11 dataset that was used to train both EfficientDet model with EfficientNet-v2 backbone and EfficientNet-B7+SVM. The dataset has 11,000 microscopic training images from 11 categories. Our results show robust performance with an accuracy of 92%, and an F1 score of 93%. Additionally, the IOU distribution illustrates the high localization capability of the detector.

## 1. INTRODUCTION

Intestinal parasitic infections (IPIs) caused by protozoan and helminth parasites are among the most common infections in humans in low-and-middle-income countries (LMICs) [1]. IPIs affect approximately 3.5 billion people, of whom the majority are children [2]. According to the WHO [3], about one-fourth of the global population is infected with intestinal parasites, primarily *Ascaris lumbricoides*, *Trichuris trichiura*, and hookworms. Furthermore, soil-transmitted helminths affect around two billion people worldwide [1]. IPIs are regarded as a severe public health concern, as they cause a wide array of potentially detrimental conditions, such as iron deficiency, growth retardation, asymptomatic carriage, diarrhea, abdominal pain, general malaise, and weakness [1], [4]–[6]. IPIs not only affect the health sector, but also the economic status of a country. These infections are among Neglected Tropical Diseases, which impact worker productivity and cause the loss of billions of dollars annually, keeping low-income countries in poverty [7].

The detection of parasitic diseases can be done through various tests, such as fecal examination, endoscopy / colonoscopy, blood tests, and radiography [8]. The diagnostic sensitivity of these techniques is low to moderate, which may be attributed to differences in sample collection or laboratory processing [9]. Diagnosis of intestinal parasites are usually based on direct examination by experienced and skilled medical laboratory technologists. Misinterpretation of laboratory analyses may occur if performed by professionals with little experience [9].

Over the last decade, researchers have been developing digital image processing models and pattern recognition techniques for the automatic identification of parasite eggs in microscopic images. The main objective of these works is to reduce human error during the diagnosis of fecal parasites, and to generate faster, more efficient, and accurate results. Due to the well-characterized and reasonably homogeneous morphology of intestinal parasites, these organisms are considered to be of great interest in the implementation of algorithms for automated recognition based on diagnostic image processing [10]. With regard to efficiency of sample processing, automated diagnostics offers high precision in the recognition of host positivity and parasite load [11]. With this, low sensitivity and time-consuming standard methods can be replaced. Artificial neural networks (ANNs) have the capability to provide automated means of identification [9], [12]– [15], thereby reducing misidentification due to human error.

## 2. RELATED WORKS

[16] proposed a CNN-based technique that used a transfer learning (TL) strategy to enhance the efficiency of automatic parasite classification in poor-quality microscopic images. This study exploited the benefits of a low-cost USB microscope, while overcoming the challenge posed by the lack of high-quality microscopes and thereby increasing

applicability and usability. [17] developed the helminth egg analysis platform (HEAP), a user-friendly microscopic helminth egg identification and quantification platform, to assist laboratory technicians during examination. This integrated multiple deep learning (DL) strategies, including a Single Shot Detector (SSD), U-Net, and Faster R-CNN, to identify the same specimen and allow users to select the best predictions. A DL approach using a CNN architecture for parasite egg classification was done by [18]. Its goal was to classify three types of *A. lumbricoides* eggs with the optimal DL architecture. A different approach was employed by [19], where a YOLOv5 algorithm was used to automatically recognize and classify parasite eggs in microscopic images of human feces.

An enhanced K-means clustering (EKM) algorithm on the modified global contrast stretching (MGCS) and modified linear contrast stretching (MLCS) enhancement techniques was proposed by [20] to analyze the segmentation performance of unsupervised color-image segmentation of helminth parasites. Moreover, [21] investigated the application of state-of-the-art (SOTA) object detectors for automatically locating and identifying parasite eggs in microscopic images. The three DL object detectors utilized were Faster R-CNN, RetinaNet, and CenterNet.

The key contributions of this paper are as follows:

- A DL-based object detection approach was utilized to train the detector to localize and classify eggs of parasitic worms using a SOTA object detector, i.e., EfficientDet and a large parasitic egg dataset.
- Comparisons were done between the proposed multi-modal learning detector and baseline methods to highlight improvements in localization and classification.

This paper was structured as follows: Section 3 describes the dataset and illustrates the proposed solution; in Section 4, the experimental setup, and results are discussed; Section 5 summarizes the significance of this work and opens doors for further improvement.

## 3. MATERIALS AND METHODS

This section describes the dataset and discusses the methods utilized in this work.

### 3.1. Datasets Overview

An ICIP2022 challenge proposed a novel dataset called Chula-ParasiteEgg-11 [22], [23] that contains various types of parasitic eggs. The dataset is composed of 11 categories of parasitic ova from stool smear samples, with average sizes ranging between 15 and 100 µm. The dataset is the largest collection of its kind, composed of 1000 and 250 microscope images for the training and testing sets, respectively, for every category. The categories include the following: *A. lumbricoides*, *Capillaria philippinensis*, *Enterobius vermicularis*, *Fasciolopsis buski*, Hookworm egg, *Hymenolepis diminuta*, *H. nana*, *Opisthorchis viverrine*, *Paragonimus* spp., *Taenia* spp. egg, and *T. trichiura*. Fig. 1 shows several examples of microscopic images with various types of parasitic eggs. The labels of the data are in the form of bounding boxes. The microscope images were acquired from several devices, including a Canon EOS 70D camera body with Olympus BX53 microscopes, a DS-Fi2 Nikon camera body with Nikon Eclipse Ni microscopes, Samsung Galaxy J7 Prime phone, and iPhone 12 and 13 with either 10× eyepiece lenses of Nikon Eclipse Ni or Olympus BX53 devices. Due to the variety of devices used, images vary in resolution, lighting, and setting conditions. Some images also exhibit noise and motion blur, having been captured with a motorized stage microscope. The large range of image characteristics could enhance detection robustness.

### 3.2. The proposed Method

#### 3.2.1 EfficientNet Algorithm

Initially, training an object identification model involves conversion of the image into properties that can serve as inputs to a NN. Significant development has been achieved in the field of computer vision by utilizing CNNs in extracting trainable characteristics from images [24]. CNNs combine and pool image information at several granularities, providing the model with a variety of potential configurations to focus on while learning image identification tasks. EfficientNet is a family of models that was designed as a new baseline network based on a scaled-up version of CNNs. EfficientNet is the foundation of the EfficientDet framework

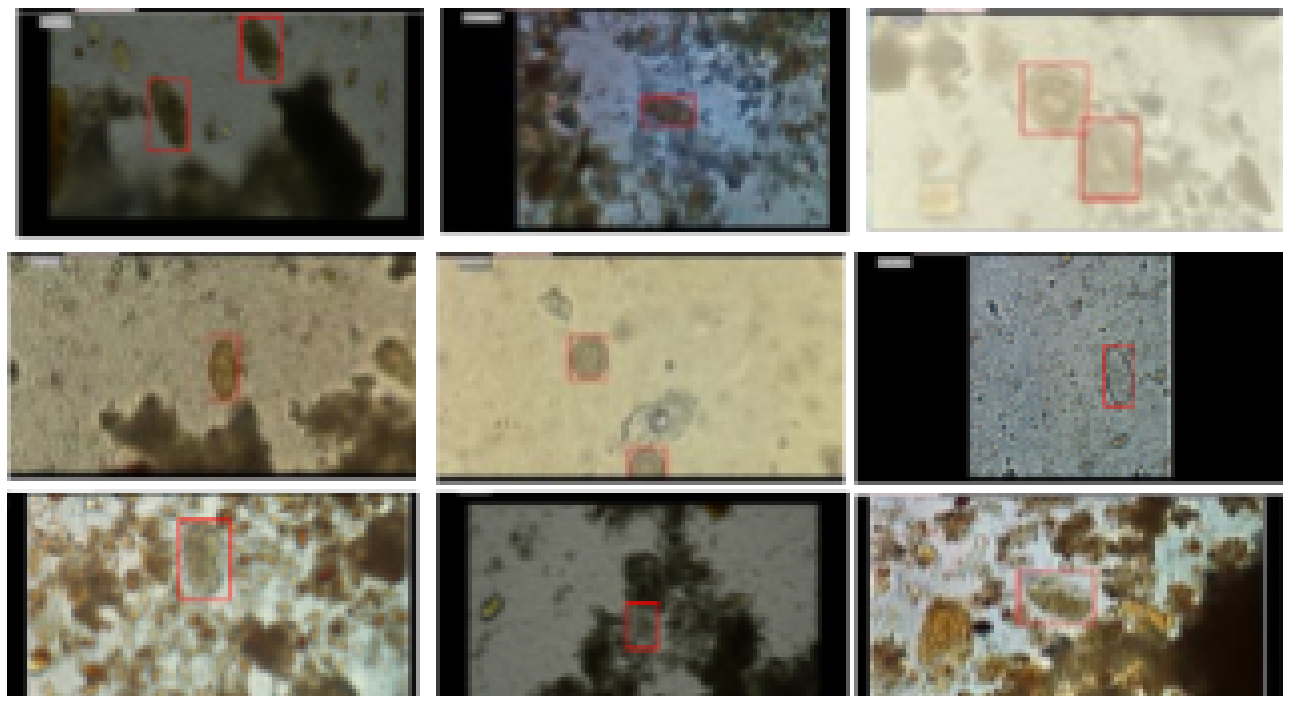

**Fig. 1:** Image examples with various resolutions, lighting, and setting conditions from the Chula-ParasiteEgg-11 dataset [22], [23].

[25]. Just as EfficientNet tackles the problem of efficient classification, EfficientDet does the same, but for detection and segmentation. CNN augmentation can be done through various techniques, and this can be done through additional parameters as shown in Fig. 2. Scaling of models can be done by increasing the width of each layer, the depth of the layers, the resolution of the images, or a combination of these methods. EfficientNet aims to scale CNN structures automatically [26].

It is the goal of feature fusion to merge samples of images that are captured at various resolutions. Traditionally, fusion employs several final feature layers from the CNN, although the specific NN used may differ. FPN is a standard method for fusing features in a top-down direction [25].

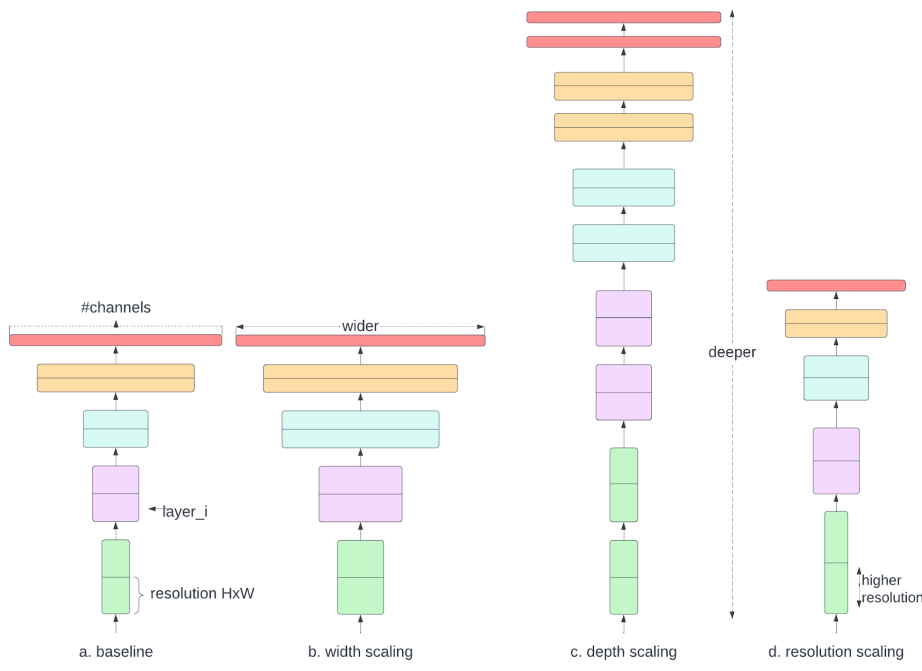

**Fig 2.** Scaling of Models. (a) baseline network; (b) — (d) conventional scaling, only scales one dimension of the network: width, depth, or resolution, respectively [26].

PANet enables reverse and forward flow of feature fusion from a lower to higher resolution [25]. On the other hand, NAS-FPN is developed from a neural architecture search and is composed of a combination of both top-down and bottom-up connections to fuse features across scales [25].

Scaling was done to dynamically resize the backbone, BiFPN, class/box, and input image quality. The network structure scales automatically with EfficientNet- B0 to EfficientNetB6. Thus, the amount of BiFPN stacks affects the network depth and breadth [25]. The EfficientDet architecture is built on the EfficientNet backbone [26], the BiFPN feature network, and a common class/box predictive network. Both BiFPN and class/box net layers are reiterated numerous times to account for resource restrictions of varying magnitude [25].

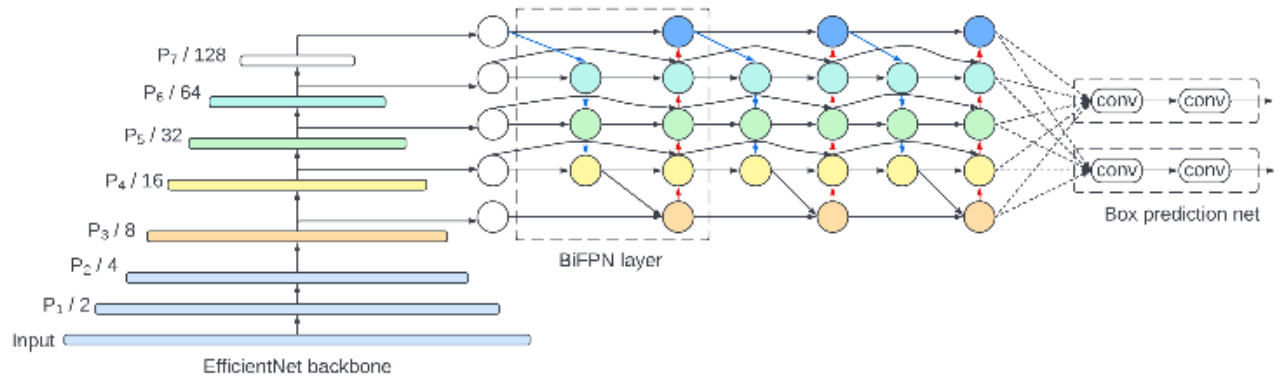

**Fig 3.** EfficientDet Architecture [25].

### 3.2.2 The proposed multi-modal learning solution

In this paper, the EfficientDet object detector with EfficientNet-v2 backbone was used to localize and classify parasitic eggs. This detector was trained for 20 epochs using the dataset of microscopic images. The detector was able to localize the eggs well while yielding high IOU values but was unable to recognize *H. diminuta* eggs. Therefore, a pre-trained EfficientNet-B7 CNN was utilized after replacing fully connected layers by a support vector machine (SVM) to enhance classification performance. The TL approach was carried out by training EfficieneNet-B7 with a large-scale dataset, i.e., ImageNet [27] and by using it with a novel medium-scale dataset, i.e., as Chula-ParasiteEgg-11 [22], [23]. This CNN was able to recognize *H. diminuta* eggs better than the detector. Finally, a multi-modal learning method that combined an EfficientDet object detector with EfficientNet-v2 backbone and EfficientNet-B7+SVM model was proposed with average decision to yield the decision about the final object category. The block diagram of the proposed solution is shown in Fig. 4.

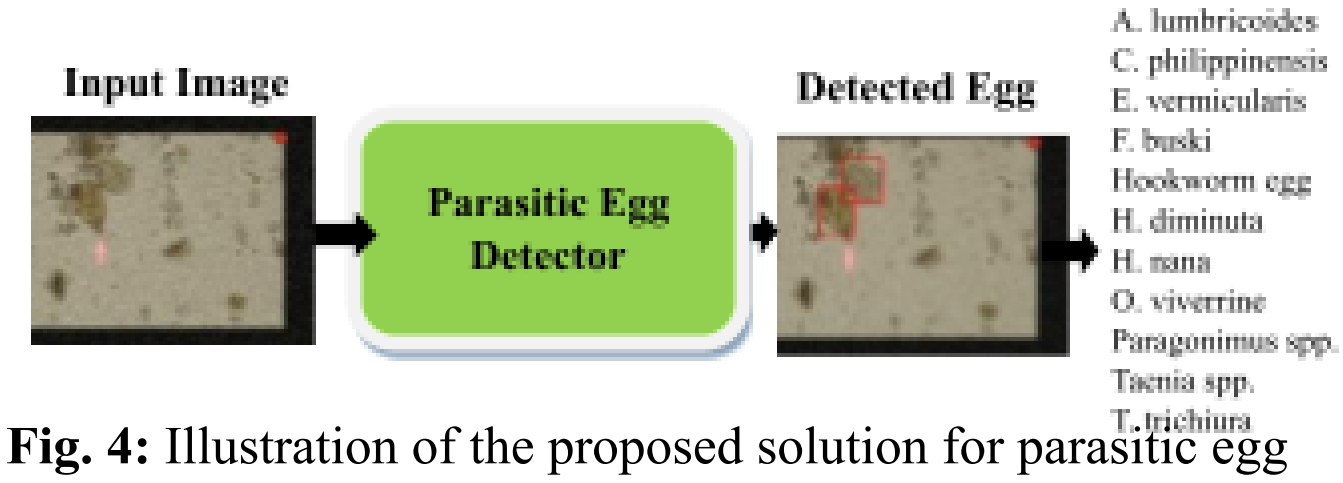

**Fig. 4:** Illustration of the proposed solution for parasitic egg detection and classification.

## 4. RESULTS AND DISCUSSION

### 4.1. Experimental Setup

The training of the proposed solution was carried out by using a PyTorch framework on an NVIDIA Tesla V100 GPU. The microscopic images were resized to 512×512 before being inputted into the EfficientDet detector. Additionally, the images were normalized using the mean and standard deviation of ImageNet. The number of epochs was set to 20. The batch size was 4. The learning rate was 0.0002. For EfficientNet-B7, to extract 2560 features, the microscopic images were resized to 600×600. To find the best accuracy for the SVM, regularization parameters and kernel functions were tuned. The $C$ was set to 5, and the $f$ was RBF.

A training set with 11000 labeled microscopic images was provided. We divided the set into three: 60% (6600 images) for training, 20% for validation (2200), and 20% for testing (2200). The results shown in Tables 1, 2 and in Fig. 7 are results of the testing dataset.

### 4.2. Experimental Results

In this section, we evaluate the performance of the proposed solution and compare it with baseline methods. First, to evaluate the detection model, an IOU distribution between actual and predicted bounding boxes was plotted in Fig. 5. It is seen that most IOUs were high, i.e., IOU > 0.8. Additionally, only 17 of 2200 images were mis-detected during the detection stage. Moreover, to evaluate the classification model, a distribution of confidence scores was plotted in Fig. 6. The detector yielded high confidence scores, i.e. confidence scores > 0.8.

To measure classification performance and compare it with baseline methods, the average accuracy of 11 categories of parasitic eggs was considered as the evaluation metric to rank the best solutions. Additionally, the average F1 score, which is the harmonic mean of precision and recall, was the second metric for ranking solutions. The recall, precision, and F1 score of the proposed solution, which uses a combination of an EfficientDet object detector with EfficientNet-v2 backbone and an EfficientNet-B7+SVM model for each egg category is shown in Table 1.

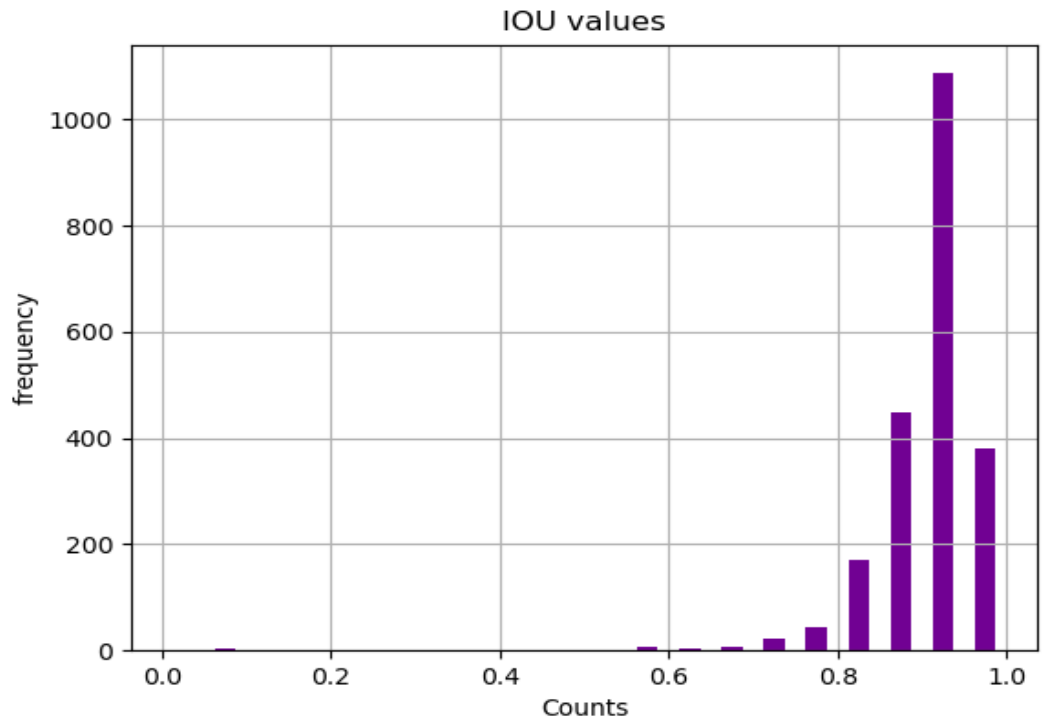

**Fig. 5:** IOU Distribution of the parasitic egg detector

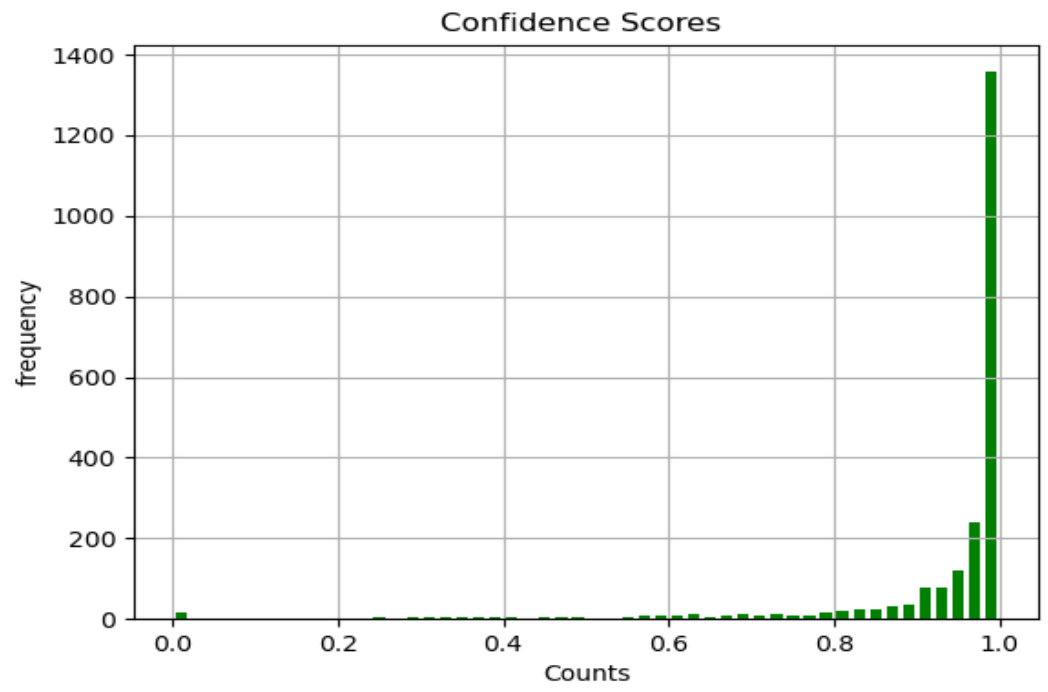

**Fig. 6:** Confidence Score Distribution of parasitic egg detector

The proposed solution was compared with baseline methods in terms of accuracy, precision, recall, and F1 score as shown in Table 2. It is evident that the proposed solution produced the highest accuracy of 92% and F1 score of 93%.

Fig. 7 shows the confusion matrix of the proposed solution including a combination of EfficientDet object detector with EfficientNet-v2 backbone and EfficientNet-B7+SVM model. The high values of the elements in the main diagonal are clear.

The confusion matrix illustrates that there was some misclassification between two types of eggs: category 5 (*H. diminuta*) and category 8 (*Paragonimus* spp.). Moreover, category 11 refers to the number of images where the detector failed to detect any eggs, i.e., only 17 of 2200 images.

## 5. CONCLUSION AND FUTURE WORK

In this paper, an efficient, fast, and highly accurate solution for parasitic egg localization and classification was proposed. A multi-modal deep learning method, including a combination of an EfficientDet object detector with EfficientNet-v2 backbone and an EfficientNet-B7+SVM model, was evaluated using a novel microscopic image dataset with 11 types of parasitic eggs.

**Table 1:** Classification Metrics of the proposed solution for each category.

| Category/Metrics | Recall | Precision | F1-score |
|---|---|---|---|
| *Ascaris lumbricoides* | 0.96 | 0.88 | 0.92 |
| *Capillaria philippinensis* | 0.95 | 0.98 | 0.97 |
| *Enterobius vermicularis* | 0.96 | 1.00 | 0.98 |
| *Fasciolopsis buski* | 0.89 | 0.95 | 0.92 |
| Hookworm egg | 0.99 | 0.99 | 0.99 |
| *Hymenolepis diminuta* | 0.66 | 0.76 | 0.70 |
| *Hymenolepis nana* | 0.98 | 0.94 | 0.96 |
| *Opisthorchis viverrine* | 0.99 | 0.97 | 0.98 |
| *Paragonimus* spp. | 0.85 | 0.71 | 0.77 |
| *Taenia spp.* egg | 0.99 | 0.98 | 0.99 |
| *Trichuris trichiura* | 0.98 | 0.99 | 0.98 |

**Table 2:** Comparison between the proposed solution and baseline methods

| Methods/ Metrics | Average Accuracy % | Average Precision % | Average Recall % | Average F1-score % |
|---|---|---|---|---|
| **EfficientNet-v2 (baseline)** | 88 | 83 | 88 | 85 |
| **EfficientNetB7+ SVM (baseline)** | 82 | 83 | 82 | 82 |
| **Multi-modal learning (proposed)** | **92** | **93** | **93** | **93** |

The results show good performance of the proposed solution which was found to outperform baseline methods in terms of accuracy, and F1 score by 4 %, and 8 % respectively. Additionally, the distribution of Intersection Over Union (IOU) illustrates high localization capability of the detector. Hence, we intend to enhance the performance of the solution by increasing the image size at input of detector and changing the detector's backbone, but this step requires more resources such as bigger RAM memory. Additionally, augmentation of microscopic images by adding noise and blurring may contribute to enhance the detection performance.

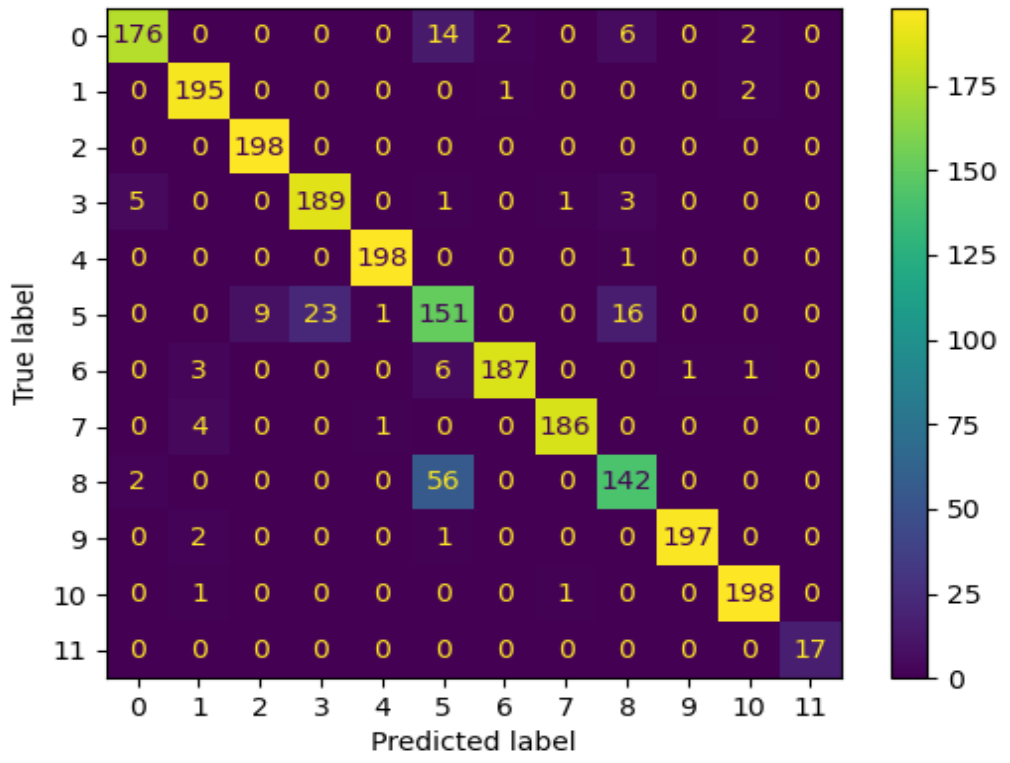

**Fig. 7:** Confusion matrix of the proposed solution for eleven categories of parasitic eggs.

## 6. ACKNOWLEDGMENT

This research project was funded by Multimedia University, Malaysia.